\documentclass{article}

\PassOptionsToPackage{numbers, compress}{natbib}
\usepackage[final]{neurips_2021}

\usepackage[utf8]{inputenc} % allow utf-8 input
\usepackage[T1]{fontenc}    % use 8-bit T1 fonts
\usepackage{hyperref}       % hyperlinks
\usepackage{url}            % simple URL typesetting
\usepackage{booktabs}       % professional-quality tables
\usepackage{amsfonts}       % blackboard math symbols
\usepackage{nicefrac}       % compact symbols for 1/2, etc.
\usepackage{microtype}      % microtypography
\usepackage{xcolor}         % colors
\usepackage{bm}
\usepackage{natbib}
\usepackage{subfig}
\usepackage{graphicx}
\usepackage{amsmath}
\usepackage{wrapfig}
\usepackage{multirow}
\usepackage{makecell}

\newcommand{\printfnsymbol}[1]{%
  \textsuperscript{\@fnsymbol{#1}}%
}

\DeclareMathOperator*{\argmax}{arg\,max}

\usepackage[ruled,vlined]{algorithm2e}
\SetKwInput{KwInput}{Input}                % Set the Input
\SetKwInput{KwOutput}{Output}               % set the Output

\title{Understanding and Improving Early Stopping for Learning with Noisy Labels}
\author{%
  Yingbin Bai$^{1}$\thanks{co-first author} \quad
  Erkun Yang$^2$\footnotemark[1] \quad
  Bo Han$^3$ \quad
  Yanhua Yang$^2$ \quad \\
  \textbf{ Jiatong Li$^4$ \quad
  Yinian Mao$^4$ \quad
  Gang Niu$^5$ \quad
  Tongliang Liu$^{1}$\thanks{Correspondence to Tongliang Liu (tongliang.liu@sydney.edu.au)}}\\[1ex]
  $^1$TML Lab, University of Sydney;
  $^2$Xidian University;
  $^3$Hong Kong Baptist University;\\
  $^4$Meituan-Dianping Group;
  $^5$RIKEN AIP
}

\begin{document}

\maketitle

\begin{abstract}
The memorization effect of deep neural network (DNN) plays a pivotal role in many state-of-the-art label-noise learning methods. To exploit this property, the early stopping trick, which stops the optimization at the early stage of training, is usually adopted. Current methods generally decide the early stopping point by considering a DNN as a whole. However, a DNN can be considered as a composition of a series of layers, and we find that the latter layers in a DNN are much more sensitive to label noise, while their former counterparts are quite robust. Therefore, selecting a stopping point for the whole network may make different DNN layers antagonistically affect each other, thus degrading the final performance. In this paper, we propose to separate a DNN into different parts and progressively train them to address this problem. Instead of the early stopping  which trains a whole DNN all at once, we initially train former DNN layers by optimizing the DNN with a relatively large number of epochs. During training, we progressively train the latter DNN layers by using a smaller number of epochs with the preceding layers fixed to counteract the impact of noisy labels. We term the proposed method as progressive early stopping (PES). Despite its simplicity, compared with the traditional early stopping, PES can help to obtain more promising and stable results. Furthermore, by combining PES with existing approaches on noisy label training, we achieve state-of-the-art performance on image classification benchmarks. The code is made public at \underline{\url{https://github.com/tmllab/PES}}.
\end{abstract}

\section{Introduction}

Deep networks have revolutionized a wide variety of tasks, such as image processing, speech recognition, and language modeling~\cite{goodfellow2016deep}, However, this highly relies on the availability of large annotated data, which may not be feasible in practice. Instead, many large datasets with lower quality annotations are collected from online queries~\cite{cha2012social} or social-network tagging~\cite{liu2011noise}. Such annotations inevitably contain mistakes or \emph{label noise}. As deep networks have large model capacities, they can easily memorize and eventually overfit the noisy labels, leading to poor generalization performance~\cite{zhang2016understanding}. Therefore, it is of great importance to develop a methodology that is robust to noisy annotations.

% \begin{figure}[!t]
% \centering
% \subfloat[features with clean labels.\label{fig:1c}]{\includegraphics[trim={0 0 0 30},width=0.4\textwidth, clip]{images/Pair20.png}}
% \subfloat[features with noisy labels.\label{fig:1d}]{\includegraphics[trim={0 0 0 30},width=0.4\textwidth, clip]{images/Pair20_noise.png}}\ 
% % {\includegraphics[width=4cm, height=3.5cm]{images/Pair200_noise.png}}     
% % {\includegraphics[width=4cm, height=3.5cm]{images/Pair200.png}}
% \caption{ t-distributed stochastic neighbor embedding (t-SNE) visualization for features on CIFAR10 with $45\%$ asymmetric noise: (a) features colored with clean labels; (b) features colored with predicted labels. We can observe that, even trained with noisy labels, the models can output discriminative features, while the classifier is severely affected by the noisy labels. }% From (a) explains the wrong decision boundary learnt by noisy labels. Both of images are generated with the same features, which extract from training examples, using a model trained in CIFAR10 45\% asymmetric label noise. Left image are colored by noisy labels, while right image is colored by clean labels. We can see that although generated features from training examples are pretty well, with each class of examples into a collection, decision boundary is misled with noisy labels, which results in fitting noisy labels well, but get poor performance with clean labels.}
% \label{fig:pair_tsne}
% \vspace{-20px}
% \end{figure}

\begin{figure}[!t]
\centering
\subfloat[Symmetric 50\%]{{\includegraphics[width=0.33\textwidth]{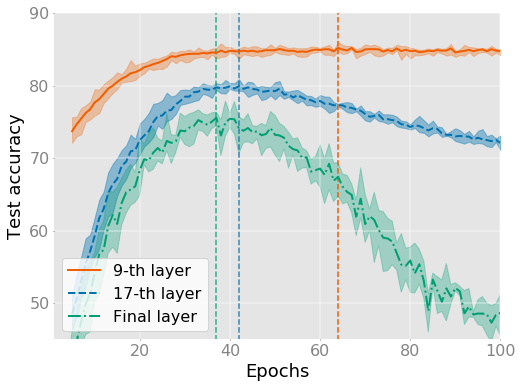}}}
\subfloat[Pairflip 45\%]{{\includegraphics[width=0.33\textwidth]{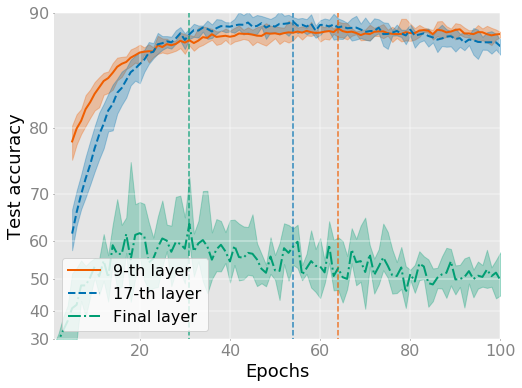}}}
\subfloat[Instance 40\%]{{\includegraphics[width=0.33\textwidth]{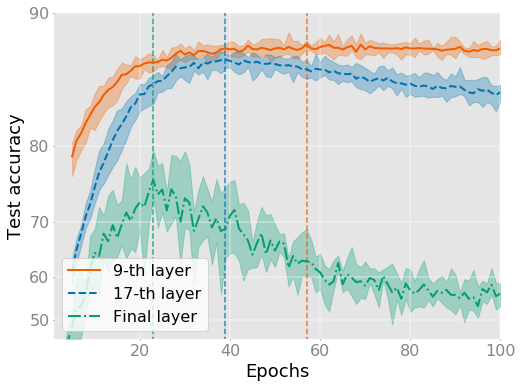}}}
\caption{We train a ResNet-18 model on CIFAR-10 with three types of noisy labels and evaluate the impact of noisy labels on the representations from the $9$-th layer, the $17$-th layer, and the final layer. The X-axis is the number of epochs for the first block of the network. The curves present the mean of five runs and the best performances are indicated with dotted vertical lines.} %\textcolor{red}{from $1$-st to $9$-th as Part 1, from $1$-st to $17$-th layer as Part 2, and the whole model as Part 3, and evaluate the quality of Part 1 and Part 2 with clean labels. The orange curve shows the quality of the whole network with noisy labels. The best performances are indicated with dotted vertical lines.}} %We can observe that the latter layers have more negative impacts from noisy labels, with earlier and larger decreases.}
\label{fig:1}
\vspace{-20px}
\end{figure}

%We can see that the impact of noisy labels for the former DNN parts is much less and later than the later DNN parts in all cases.} %The lines demonstrate the changes of quality of representations from different layers. The left image trains with symmetric label noise, and the right one is from the asymmetric label noise.  The green line shows the original performance of classifier training with label noise. For red and blue lines,  we evaluate the quality of representations by fixing the below layers and using clean labels to re-train above layers for 100 epochs. The dash lines indicate the epoch with the highest quality of representations during the training process.}

Existing methods on learning with noisy labels (LNL) can be mainly categorized into two groups: model-based and model-free algorithms.
Methods in the first category mainly model noisy labels with the noise transition matrix~\cite{Patrini2017forward, Xu2019DMI, Xie2020UDA, xia2020part}. With perfectly estimated noise transition matrix, models trained with corrected losses can approximate to the models trained with clean labels. However, current methods are usually fragile to estimate the noise transition matrix for heavy noisy data and are also hard to handle a large number of 
classes~\cite{han2018co}. The second type explores the dynamic process of optimization policies, which relates to the memorization effect$-$%if clean labels are of majority within the noisy labels for each class, 
deep neural networks tend to first memorize and fit majority (clean) patterns and then overfit minority (noisy) patterns~\cite{Arpit2017Look}. 
%\footnote{We can explicitly say that, for learning with noisy labels, we have the assumption that clean labels are of majority within the noisy labels for each class.}.
Recently, based on this phenomenon, many methods~\cite{han2018co, Wang2018Iterative, Li2020DivideMix, Liu2020ELR, xia2021robust} have been proposed and achieved promising performance.

%\footnote{We may set co-teaching and dividemix as examples here.}.% . To leverage this memorization effect, current methods usually employ the early stopping trick  to filter our noisy instances at the early stage of training and have achieved state-of-the-art performance.

% \begin{figure}[!ht]
% \centering
% {\includegraphics[trim={0 0 0 30},width=0.95\textwidth, clip]{images/clothing1m_lrp_2_3_20.png}}
% \caption{The group of images is generated by LRP technique with \textit{VGG} network \cite{DBLP:Karen2015VGG}, and the original image comes from \textit{Clothing1M} dataset \cite{Xiao2015Clothing}. These fourteen images correspond to fourteen outputs of networks. From the images, the representations of images are learned well e.g. collars, edges of clothes and girl's head. However, the results of the second and fourth images in the first row are very close, which suggests the classifier (the final layer) is confused with each other category due to the asymmetric label noise}
% \label{fig:lrp}
% \end{figure}

%\footnote{the caption of Figure 1 needs improvement. It is unclear.}

To exploit the memorization effect, when the double descent phenomenon \cite{belkin2019reconciling,nakkiran2020deep,ishida2020we} cannot be guaranteed to occur, a core issue is to study when to stop the optimization of the network. While stopping the training for too few epochs can avoid overfitting to noisy labels, it can also make the network underfit to clean labels. Current methods~\cite{tanaka2018joint, Nguyen2020SELF} usually adopt an early stopping strategy, which decides the stopping point by considering the network as a whole. However, since DNNs are usually optimized with stochastic gradient descent (SGD) with backpropagation, supervisory signals will gradually propagate through the whole network from latter layers (i.e., layers that are closer to output layers) to former layers (i.e., layers that are closer to input layers).  
Noting that the output layer is followed by the empirical risk in the optimization procedure. We hypothesize that noisy labels may have more severe impacts for the latter layers, which is different from current methods~\cite{han2018co, Li2020DivideMix} that usually stop the training of the whole network at once.

%Recent biological research~\cite{majaj2015simple} shows that, when human face indistinguishable instances, the inferior temporal (IT) neurons, which is near the end of the visual information flow, will fire, while the earlier stops along with the information flow, which is believed to process basic visual elements such as brightness and orientation, are less affected. Furthermore, since DNNs are usually trained with stochastic gradient descent and back propagation, the latter layers, which are closer to noisy labels may be probably more sensitive to noisy labels. %  With the understanding that visual images are processed through a DNN with multiple DNN layers, and early DNN layers are considered to capture basic features such as color and direction, 
%Consequently, selecting a single stopping point for the whole network may not be optimal.

To empirically verify the above hypothesis, we analyze the impact of noisy labels on representations from different layers with different training epochs. To quantitatively measure the impact of noisy labels from intermediate layers, we first train the whole network on noisy data with different training epochs and fix the parameters for the selected layer and its previous layers. We then  reinitialize and optimize the rest layers with clean data, and the 
final classification performance is adopted to evaluate the impact of noisy labels. For the final layer, we directly report the overall classification performance. As illustrated in Figure~\ref{fig:1}, we can see that latter layers always achieve the best performance at relatively smaller epoch numbers and then exhibit stronger performance drops with additional training epochs, which verifies the hypothesis that noisy data may have more severe impacts for latter layers. 
 With this understanding, we can infer that the early stopping, which optimizes the network all at once, may fail to fully exploit the memorization effect and induce sub-optimal performance.
 
%the performance of part 1 always begin to decrease while the performance of part 2 and 3 can be continually improved, and part 2 also achieves the best performance earlier than part 3.  achieves the  the impact of noisy labels for the former DNN parts is much less and later than the later DNN parts, 
%Since different DNN layers cannot be considered as a whole, 
%since different DNN layers achieve their best performance with different training epochs. %Therefore, the currently early stopping trick may fail to fully exploit the memorization effect and induce sub-optimal performance.  % and may adversary  the model performance and inducing large performance variance.
% \footnote{figure 1 is too big and looks non-professional. What do you meaning by asymmetric label noise? Pairflip? You may consider add a figure for instance-dependent label noise} \footnote{the paragraph is too long. We can divide it into several paragraph to make the logic more clear.}

To address the above problem, we propose to optimize a DNN by considering it as a composition of several DNN parts and present a novel progressive early stopping (PES) method.  Specifically, we initially train former DNN layers by optimizing them with a relatively large number of epochs. Then, to alleviate the impact of noisy labels for latter layers, we reinitialize and progressively train latter DNN layers by using smaller numbers of epochs with preceding DNN layers fixed. Since different layers are progressively trained with different early stopping epochs, we term the proposed method as progressive early stopping (PES).  Despite its simplicity, compared with normal early stopping trick, PES can help to better exploit the memorization effect and obtain more promising and stable results. Moreover, since the model size and training epochs are gradually reduced during the optimization procedure, the training time of PES is only slightly greater than that of the normal early stopping. Finally, by combining PES with existing approaches on noisy label training tasks, we establish new state-of-the-art (SOTA) results on CIFAR-10 and CIFAR-100 with synthetic noise. We also achieve competitive results on one dataset with real-world noise: Clothing-1M~\cite{Xiao2015Clothing}.
%parts\footnote{it is not necessary to mention the technical details here. Why you want to mention "different parts"? In the introduction, we just show the philosophy, i.e., different layers have different robustness property to label noise. We would like to treat them differently in early stopping to better exploit the memorization effect.}
%\footnote{the readers will have the concern of high computational complexity of the proposed method because we train the model several rounds and looks the computational complexity is relatively high. We need to defence this in the introduction or the experimental part.}

% This progressive training can help to better exploit the memorization effect. all feature learning layers and the last FC layer separately, where the feature learning layers and the last FC layer are  optimize progressively with a two-stage learning procedure. Specifically, we first train feature learning layers by optimizing the whole network with normal early stopping trick, as the feature learning layers are less affected in this procedure. Then, we reinitialize the last FC layer and apply the early stopping trick again to optimize the parameters of the last FC layers.

The rest of the paper is organized as follows. In Section~\ref{sec:3}, we first introduce the proposed progressive early stopping and then present the details of the proposed algorithm by combining our method with existing approaches on noisy label training tasks. Section~\ref{sec:exp} shows the experimental results of our proposed method.  Related works are
briefly reviewed in Section~\ref{sec:rel}. Finally, concluding remarks are given in Section~\ref{sec:con}.

% \begin{equation}
% \begin{aligned}
%     R = \sum_{ij}{||\bm{\hat{y}_{i}} - \bm{\hat{y}_{j}}||}^{2}A_{ij}\\
%     A_{ij} = \frac{exp(-||f_i  - f_j||^{2})}{2{\sigma}^{2}}
% \end{aligned}
% \end{equation}

%%%%------------------------------------------------------

%%%%------------------------------------------------------
\section{Proposed Method}
\label{sec:3}
Let $D$ be the distribution of a pair of random variables $(\bm{X}, \bm{Y})\in \mathcal{X}\times \{1,...K\}$, where $\bm{X}$ indicates the variable of instances, $\bm{Y}$ is the variable of labels, $\mathcal{X}$ denotes the feature space, and $K$ is the number of classes. In many real-world problems, examples independently drawn from the distribution $D$ are unavailable. Before being observed, the clean labels are usually randomly corrupted into noisy labels. Let $\tilde{D}$ be the distribution of the noisy example $(\bm{X}, \bm{\tilde{Y})}$, where $\bm{\tilde{Y}}$ indicates the variable of noisy labels. For label-noise learning, we can only access a sample set $\{\bm{x}_{i}, \tilde{y}_{i})\}_{i = 1}^{n}$ independently drawn from $\tilde{D}$. The aim is to learn a robust classifier from the noisy sample set that can classify test instances accurately. 

In the following, we first elaborate on the proposed progressive early stopping (PES). Then, based on PES, we provide a learning algorithm that learns with confident examples and semi-supervised learning techniques.
%\footnote{first time in the method, the term should in its full version.}. 
%\footnote{why you want to study the two cases? Explain. You can discuss this after section 3.1} 
%\section{The Early Stopping Trick}
\subsection{Progressive Early Stopping}
When trained with noisy labels, if clean labels are of majority within each noisy class, deep networks tend to first fit clean labels during an early learning stage before eventually memorizing the wrong labels, which can be explained by the memorization effect.
%The memorization effect indicates that deep networks tend to first memorize and fit easy (clean) patterns before eventually memorizing hard (noisy) patterns. \footnote{explain this in more details.} 
Many current methods utilize this property to counteract the influence of noisy labels by stopping the optimization at an early learning phase. Specifically, a deep classifier can be obtained by optimizing the following objective function with a relatively small epoch number $T$:
\begin{equation}
    \begin{aligned}
    \min_{\Theta}\frac{1}{n}\sum_{i = 1}^{n}\mathcal{L}(f(\bm{x}_{i};\Theta), {\tilde{y}_{i}}),
    \end{aligned}
\label{eq:1}
\end{equation}
where $f(\cdot;\Theta)$ is a deep classifier with model parameters $\Theta$ and $\mathcal{L}$ is the cross-entropy loss. When trained with noisy data, early learning regularization (ELR) %\footnote{first time with a full name.}
\cite{Liu2020ELR} reveals that, for the most commonly used cross-entropy loss,  the gradient is well correlated with the correct direction at the early learning phase. Therefore, with a properly defined small epoch number $T$, the classifier can have higher accuracy than at initialization. While, if we continue to optimize the deep model after $T$ epochs, the classifier will be able to memorize more noise labels. Therefore, it is critical to select a proper epoch number $T$ to utilize the memorization effect and alleviate the influence of noisy labels.

Current methods typically select the epoch number $T$ by considering the network as a whole. However, as Figure \ref{fig:1} makes clear, the impact of noisy labels on different DNN layers are different, which implies that the traditional early stopping trick, which optimizes the whole network all at once, may make different DNN layers to be antagonistically affected by each other, thus degrading the final model performance. 
\begin{figure}[!t]
\centering
% \subfloat[features with clean labels.\label{fig:1a}]{\includegraphics[trim={0 0 0 30},width=0.3\textwidth, clip]
\subfloat[Symmetric 50\%]{\includegraphics[width=0.33\textwidth]{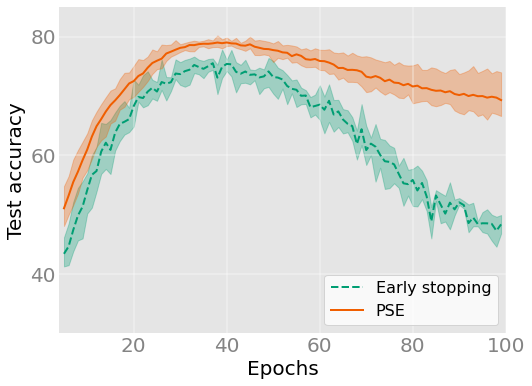}} 
\subfloat[Pairflip 45\%]{\includegraphics[width=0.33\textwidth]{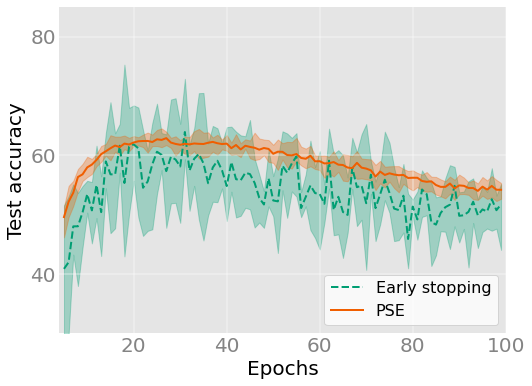}} 
\subfloat[Instance 40\%]{\includegraphics[width=0.33\textwidth]{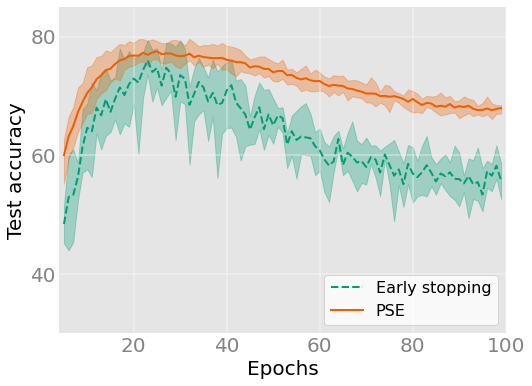}}
\caption{Performance of the traditional early stopping trick and the proposed PES on CIFAR-10 with different types of label noise. The lines present the mean of five runs.}
% (a) 50\% symmetric noise; (b) 45\% pair noise; (c) 40\% instance-dependent noise. 
\label{fig:2}
\vspace{-10px}
\end{figure}
To this end, we propose to separate a DNN into different parts and progressively train layers in different parts with different training epochs. Specifically, assume that the whole network $f(\cdot; \Theta)$ can be constituted with $L$ DNN parts
%\footnote{philosophically, we dont need to split the neetwork into k parts. We are doing so as an implementation of our idea. We should first make our philosophy clear. Then, making it clear that spliting the network into k parts is just one implementation. It is also a good idea to introduce our idea by arguing that each layers should be treated differently. Then, discuss the k split in the implementation part.}
\begin{equation}
\begin{aligned}
    &\bm{z}_{1} = f_{1}(\bm{x}; \Theta_{1}),\\
    &\bm{z}_{l} = f_{l}(\bm{z}_{l-1}; \Theta_{l}), \quad  l = 2,\dots, L \\
   % &\hat{\bm{y}} = f_{k}(\bm{z}_{k-1}; \Theta_{k}).
    \end{aligned}
\end{equation}
where $f_{l}(\cdot; \Theta_{l})$ is the $l$-th DNN part and $\bm{z}_{l}$ is the corresponding output. The output of the last part $\bm{z}_{L}$ is the prediction. The network $f(\cdot; \Theta)$ can also be represented as $f(\cdot; \Theta_{1},...\Theta_{L})$. To counteract the impact of noisy labels, We initially optimize the parameter $\Theta_{1}$ for the first part by training the whole network for $T_{1}$ epochs with the following objective
\begin{equation}
    \begin{aligned}
    \min_{\Theta_1\dots \Theta_{k}}\frac{1}{n}\sum_{i = 1}^{n}\mathcal{L}(f(\bm{x}_{i}; \Theta_{1},\dots,\Theta_{L}), {\tilde{y}_{i}}). 
    \end{aligned}
    \label{eq:3}
\end{equation}
Then, we keep the obtained parameter $\Theta_{1}^{*}$ fixed, reinitialize and progressively learn the $l$-th ($l = 2,\dots, L$) DNN part with the parameters for preceding DNN parts fixed. The training procedure is conducted with  $T_{l}$ epochs by optimizing the following objective%\footnote{Keep $\Theta_{i}^{*}$ fixed and then progressively learn $\Theta_{i+1}^{*}$?}
\begin{equation}
    \begin{aligned}
    \min_{\Theta_l\dots \Theta_{k}}\frac{1}{n}\sum_{i = 1}^{n}\mathcal{L}(f(\bm{x}_{i}; \Theta_{1}^{*},\dots,\Theta_{l-1}^{*},\Theta_{l},\dots, \Theta_{L}), {\tilde{y}_{i}}), \quad  l = 2\dots L
    \end{aligned}
    \label{eq:4}
\end{equation}
We gradually optimize the $(l+1)$-th DNN part with the obtained parameter $\Theta_{l}^{*}$ fixed, the optimization is continued until all the parameters have been optimized. As elaborated above, latter DNN parts are more sensitive to noisy labels than their former counterparts. Therefore, for the above initializing optimization in Eq.~\eqref{eq:3} and the following $L-1$ steps of optimization in Eq.~\eqref{eq:4}, we gradually reduce the training epochs (i.e. $T_1 \geq T_2 \geq \dots\geq T_L$) to better exploit the memorization effect. 
After optimization, we can obtain the final network as $f(\cdot, \Theta) = f(\cdot; \Theta_{1}^{*},\dots,\Theta_{L}^{*})$. 
Since this model is obtained by progressively exploiting the early stopping strategies for different DNN parts, we term the proposed method as progressive early stopping (PES). %also apply the normal early stopping trick to counteract the impact of noisy data for the corresponding DNN part\footnote{unclear. Rewrite}. 

To explicitly verify the effectiveness of the proposed PES method, we conduct several pilot experiments, which compare the traditional early stopping and PES with label noise from different types and different levels. The results are illustrated in Figure~\ref{fig:2}, from which we can see that, compared with models trained with traditional early stopping,  models trained with PES can achieve superior classification accuracy with smaller variations in all the cases. Current state-of-the-art methods~\cite{Li2020DivideMix} usually adopt models with the traditional early stopping as base models to distill confident examples and then utilize semi-supervised learning techniques by considering confident examples as labeled data and other noisy examples as unlabeled data to further improve the results. The final performance still heavily relies on the base model trained with noisy labels. By improving the performance of the base model, our method combined with semi-supervised learning techniques is able to establish new state-of-the-art results. In the following subsections, we will elaborate on how to utilize PES to distill confident examples and further combine it with semi-supervised learning techniques.

\subsection{Learning with Confident Examples}
Based on the deep network optimized with progressive early stopping, we can select confident examples to facilitate the model training. Here, confident examples refer to examples that have high probabilities with clean labels. In this paper, we treat examples whose predictions are consistent with given labels as confident examples. In addition, to make the results more robust,  we generate two different augmentations for any given input and use the average prediction to decide its predicted label.  Formally, we can obtain the confident example set $\mathcal{D}_{l}$ as
%one training example $(\bm{x},\bm{\tilde{y}})$ is regarded as confident example if

\begin{equation}
\label{eq:5}
    \begin{aligned}
    \begin{gathered}
 \mathcal{D}_{l} = \{(\bm{x}_{i},{\tilde{y}}_{i})|{\tilde{y}}_{i} = \hat{{y}}_{i}, i = 1,\dots,n\},\\
 {\hat{y}}_{i} = \argmax_{k\in\{1,\dots,K\}}\frac{1}{2}[f^{k}(\text{Augment}(\bm{x}_{i});\Theta) + f^{k}(\text{Augment}(\bm{x}_{i});\Theta)],
     \end{gathered}
    \end{aligned}
\end{equation}

%$(\bm{x},\bm{\tilde{y}})$,we treat the  For confident examples selection, the classifier will guess two times of each example with different image augmentations, and use the class with the sum maximum probability in two guesses, regarding ones that are identical with their noisy labels as confident examples. In the latter training process, our framework will train the classifier by alternately update of network parameters and confident examples every epoch. Algorithm \ref{algorithm} provides detailed pseudo code for our framework. 

% \begin{equation} 
% (X_i, Y_i) \in ({X}', {Y}') \ \ \ if \ \ \hat{P}_i = Y_i 
% \end{equation}

% \begin{equation} 
% \hat{P} = argmax(\frac{1}{2} (P_{model}(y| Aug_1(x) ; \theta) + P_{model}(y| Aug_2(x) ; \theta)))
% \end{equation}

where $\text{Augment}(\cdot)$ indicates normal data augmentation operation including horizontal random flip and random crops, and $f^{k}(x;\Theta)$ is the predicted probability of $\bm{x}$ belonging to class $k$. Note that $\text{Augment}(\cdot)$ is a stochastic transformation, so the two terms in Eq \eqref{eq:5} are not identical. The average prediction of augmented examples provides a more stable prediction and is found empirically to improve performance. After obtaining the confident example set, one can easily train a classifier by considering confident examples as clean data. However, since the number of confident examples for different classes can vary greatly, directly training the model with the obtained confident example set may introduce a severe class imbalance problem. To this end, we adopt a weighted classification loss %The corresponding algorithm is summarized in Algorithm~\ref{}.

\begin{equation}
\label{eq:6}
\begin{aligned}
\mathcal{L}_{c} = \sum_{i=1}^{N}w_{y_{i}}\mathcal{L}_{p}(\tilde{y}_{i}, f(\bm{x}_{i}; \Theta)),  \\
%\frac{\sum^N_{i=1}  weight[class[i]] * H(\tilde{y}, f(\hat{y}|x; \Theta))}{\sum^N_{i=1} weight[class[i]]}
\end{aligned}
\end{equation}
where $w_{i}$ is the corresponding class weight. Assuming that $\sigma_{k} = |\{(\bm{x}_{i}, \tilde{y}_{i})|\tilde{y}_{i} = k,(\bm{x}_{i}, \tilde{y}_{i})\in{\mathcal{D}_{l}}\}| $  denotes  the cardinality of the confident example  set belonging to the $k$-th class. Then, we can set $w_{i} = {\sigma_{i}}/{(\sum_{j=1}^{K}\sigma_{j})}$ to indicate the corresponding class importance.
% \begin{equation}
%     \begin{aligned}
%     w_{i} = |{(\bm{x}_{i}, \tilde{y}_{i})|\tilde{y}_{i} = i\in{\mathcal{D}_{l}}}
%     \end{aligned}
% \end{equation}
% Extracted numbers of confident examples for different categories may be seriously unbalanced, which degenerates the performance of classifiers. We employ class weights in Eq \eqref{eq:weights} to balance losses from different categories, which are calculated by inverse class frequency of extracted examples, and normalized across classes \cite{Huang16Imbalanced, Wang17LongTail}. Compared with the prior regularization used in \cite{tanaka2018joint, Li2020DivideMix}, it does not require the knowledge of the prior distribution, which means the examples in noisy datasets can be unbalance. We give details of the algorithm without semi-supervised learning in [?].

\subsection{Combining with Semi-Supervised Learning}
\label{sec:2.3}
Training with only confident examples neglects the rest data and may suffer from insufficient training examples. To tackle this problem, we further resort to semi-supervised learning techniques by considering confident examples as labeled data and other noisy examples as unlabeled data. Specifically, the labeled data set and unlabeled data set can be obtained as
\begin{equation}
\label{eq:7}
    \begin{aligned}
    \begin{gathered}
 \left\{\begin{matrix}
 \mathcal{D}_{l} = \{(\bm{x}_{i},{\tilde{y}}_{i})|{\tilde{y}}_{i} = \hat{y}_{i}, i = 1,...n\} \\
 \mathcal{D}_{u} = \{\bm{x}_{i}|{\tilde{y}}_{i} \neq \hat{y}_{i}, i = 1,...n\} \end{matrix}\right.\\
 {\hat{y}}_{i} = \argmax_{k\in\{1,...K\}}\frac{1}{2}[f^{k}(\text{Augment}(\bm{x}_{i});\Theta) + f^{k}(\text{Augment}(\bm{x}_{i});\Theta)],
     \end{gathered}
    \end{aligned}
\end{equation}
where the labeled data set $\mathcal{D}_{l}$ is the same as that in Eq \eqref{eq:6}, and $\mathcal{D}_{u}$ is the rest unlabeled data set. Similar to~\cite{Li2020DivideMix}, we adopt MixMatch~\cite{Berthelot2019MixMatch} as the semi-supervised learning framework to train the final classification models. For more details about semi-supervised learning, we refer to ~\cite{Berthelot2019MixMatch}. The whole learning algorithm is summarized in Algorithm~\ref{algorithm1}.

\begin{algorithm}[!tp]
{\bfseries Input}:  Neural network with trainable parameters $\Theta = \{\Theta_{1},\dots,\Theta_{L}\}$, Noisy training dataset $\{\bm{x}_{i}, \tilde{y}_{i})\}_{i = 1}^{n}$, Number of training epochs for different part: $T_{1},\dots,T_{L} $, and training epochs $T_{c}$ for refining with confident examples.%NumEpoch$, Number of refine epoch $NumRefine_1$, and $NumRefine_2$, Number of warmup epoch $NumWarmup$

%\tcc{Early stopping}
\For{$i = 1, \dots, T_{1}$ }{
     {Optimize network parameter} $\Theta$  with Eq.~\eqref{eq:3}; \\
}

%\tcc{PES}
\For{l = 2, \dots, L}{
 {Froze} $\{\Theta_{1},\dots,\Theta_{l-1}\}$ and re-initialize $\{\Theta_{l},\dots,\Theta_{L}\}$; \\
\For{$i = 1, \dots, T_{l}$ }{
     {Optimize network parameter} $\{\Theta_{l},\dots,\Theta_{L}\}$  with Eq.~\eqref{eq:4}; \\
}
}
Unfroze $\Theta$;\\

\For{$i = 1, \dots, T_{c}$ }{
{Extract} confident example set $\mathcal{D}_{l}$ and unlabeled set $\mathcal{D}_{u}$ with classifier $f(\cdot, \Theta)$ by Eq.~\eqref{eq:7};\\
	 Training the classifier $f(\cdot, \Theta)$ with MixMatch loss on $\mathcal{D}_{l}$ and $\mathcal{D}_{u}$;}
     {Evaluate} the obtained classifier $f(\cdot, \Theta)$. \\
\caption{Progressive Early Stopping with Semi-Supervised Learning}
\label{algorithm1}
\end{algorithm}

%%%%------------------------
\section{Experiments}
\label{sec:exp}
% In this section, we first introduce the adopted datasets and implementation details and then present and discuss the experimental results together with  comparative evaluations with other state-of-the-art baselines.
\subsection{Datasets and Implementation Details}
\label{sec:impl}
\textbf{Datasets:} We evaluate our method on two synthetic datasets, CIFAR-10 and CIFAR-100 \cite{krizhevsky2009CIFAR} with different levels of symmetric, pairflip, and instance-dependent label noise (abbreviated as instance label noise) and a real-world dataset Clothing-1M \cite{Xiao2015Clothing}. Both CIFAR-10 and CIFAR-100 contain 50k training images and 10k test images of size $32\times 32$. Following previous works~\cite{han2018co, xia2019anchor, Liu2020ELR, xia2021robust}, symmetric noise is generated by uniformly flipping labels for a percentage of the training dataset to all possible labels. Pairflip noise flips noisy labels into their adjacent class. And, instance noise is generated by image features. More details about the synthetic label noise are given in the \emph{supplementary material}. For the flipping rate, it can include \cite{han2018co, xia2019anchor} or ex-include \cite{Li2020DivideMix, Liu2020ELR} true labels. We use the flipping rate including correct labels in Table \ref{tab:cifar_sym} to compare with results in \cite{Li2020DivideMix}, and use without correct labels in the rest of the experiments. Clothing-1M \cite{Xiao2015Clothing} is a large-scale dataset with real-world noisy labels, whose images are clawed from the online shopping websites, and labels are generated based on surrounding texts. It contains 1 million training images, and 15k validation images, and 10k test images with clean labels. 

\renewcommand{\arraystretch}{1.15}
\begin{table}[!tp]
\fontsize{8.5}{10}\selectfont
\centering
\caption{Preliminary analysis of the performance and the quality of extracted confident examples on CIFAR-10. The mean and standard deviation are computed over five runs.}
\vspace{5pt}
\scalebox{1}{
\begin{tabular}{c | c | c | c | c | c | c}
\Xhline{2\arrayrulewidth}
	Metrics                         & Methods  & Sym-20\%         & Sym-50\%        & Pair-45\%       & Inst-20\%       & Inst-40\%          \\ \hline
\multirow{2}{*}{Test Accuracy} 	    & Early Stopping   & 82.55$\pm$2.46   & 70.76$\pm$1.24  & 60.62$\pm$5.59  & 84.41$\pm$0.90  & 74.73$\pm$2.65    \\ %\cline{2-7}
	                                & PES              & \textbf{85.87$\pm$1.59}   & \textbf{75.87$\pm$1.33}  & \textbf{62.40$\pm$2.34}  & \textbf{86.58$\pm$0.45}  & \textbf{77.07$\pm$1.18}    \\ \hline
\multirow{2}{*}{Label Precision} 	& Early Stopping       & 98.81$\pm$0.15   & 94.65$\pm$0.19  & 72.53$\pm$5.26  & \textbf{98.70$\pm$0.43}  & \textbf{90.77$\pm$1.87}    \\ %\cline{2-7}
	                                & PES              & \textbf{98.96$\pm$0.09}   & \textbf{95.46$\pm$0.14}  & \textbf{72.99$\pm$2.27}  & 98.52$\pm$0.19  & 90.63$\pm$0.92    \\ \hline
\multirow{2}{*}{Label Recall} 	    & Early Stopping   & 88.51$\pm$2.26   & 75.18$\pm$1.00  & 67.84$\pm$5.06  & 90.37$\pm$1.01  & 82.15$\pm$3.17    \\ %\cline{2-7}
	                                & PES              & \textbf{92.67$\pm$1.43}   & \textbf{81.03$\pm$1.83}  & \textbf{71.06$\pm$2.27}  & \textbf{93.24$\pm$0.60}  & \textbf{85.91$\pm$0.68}    \\ \hline
\Xhline{2\arrayrulewidth}
\end{tabular}
}
\label{tab:pre}
\vspace{-5pt}
\end{table}

\textbf{Baselines:} Semi-supervised learning may strongly boost the performance, we separately compare our method with approaches with or without semi-supervised learning. For the comparison with baselines with semi-supervised learning, we combine our proposed method with MixMatch used in \cite{Li2020DivideMix} as indicated in Subsection~\ref{sec:2.3}. (1) Approaches without semi-supervised learning: Co-teaching \cite{han2018co}, Forward \cite{Patrini2017forward}, Joint Optim \cite{tanaka2018joint}, T-revision \cite{xia2019anchor}, DMI \cite{Xu2019DMI}, and CDR \cite{xia2021robust}. (2) Methods with semi-supervised learning: M-correction \cite{Arazo2019UnsupervisedLabel}, DivideMix \cite{Li2020DivideMix}, and ELR+ \cite{Liu2020ELR}. We also adopt standard training with cross-entropy (CE) and MixUp \cite{Zhang2017MixUp} as baselines to show improvements.

\textbf{Network structure and optimization:} Our method is implemented by PyTorch v1.6. Baseline methods are implemented based on public codes with hyper-parameters set according to the original papers. For DivideMix and ELR+, we evaluate the test accuracy with the first network. To better demonstrate the robustness of our algorithm, we keep the hyper-parameters fixed for different types of label noise. More technique details are given in the \emph{supplementary material}.

For experiments without semi-supervised learning, we follow \cite{xia2019anchor}, and use ResNet-18 \cite{He2016ResNet} for CIFAR-10 and ResNet-34 for CIFAR-100. We split networks into three parts, the layers above block 4 as part 1, block 4 of ResNet as part 2, and the final layer as part 3. $T_1$ is defined as 25 for CIFAR-10 and 30 for CIFAR-100, $T_2$ as 7, and $T_3$ as 5. The network is trained for 200 epochs and SGD with $0.9$ momentum is used. The initial learning rate is set to $0.1$ and decayed with a factor of $10$ at the 100th and 150th epoch respectively, and a  weight decay is set to $10^{-4}$. For $T_2$ and $T_3$, we employ an Adam optimizer with a learning rate of $10^{-4}$.

For experiments with semi-supervised learning, we follow the setting of \cite{Li2020DivideMix} with PreAct Resnet-18. We set the final layer as part 2, the rest as part 1. $T_1$ is defined as 20 for CIFAR-10 and 35 for CIFAR-100, and $T_2$ as 5. The network is trained for $300$ epochs. For optimization, we use a single cycle of \textit{cosine annealing} \cite{Loshchilov2017CosineLr}, and the learning rate begins from $2 \times 10^{-2}$ and ends at $2 \times 10^{-4}$, with a weight decay of $5 \times 10^{-4}$. An Adam optimizer is adopted with a learning rate of $10^{-4}$ for $T_2$. For hyper-parameters from MixMatch, we set them according to the original paper \cite{Berthelot2019MixMatch}.

For Clothing-1M \cite{Xiao2015Clothing}, we follow the previous work \cite{tanaka2018joint}, and employ a ResNet-50 \cite{He2016ResNet} pre-trained on ImageNet~\cite{krizhevsky2012imagenet}. We set the final layer as part 2, the rest as part 1. $T_1$ and $T_2$ are defined as 20 and 7 respectively. The network is trained with CE loss for 50 epochs and SGD is used with $0.9$ momentum and a weight decay of $10^{-3}$. The learning rate is $5 \times 10^{-3}$ and decayed by a factor of $10$ at the 20th and 30th epoch respectively. We employ an Adam optimizer with a learning rate of $5 \times 10^{-6}$ for $T_2$.

% , which are usually mixed together for real-world problems.
% with a batch size of $128$

\subsection{Preliminary Experiments}
In Figure \ref{fig:2}, we can observe that with the PES trick, the performance of classifiers is generally improved compared with that the traditional early stopping trick. In this section, we further carefully analyze the quality of extracted labels by examining them from three aspects, i.e., test accuracy, label precision, and label recall. Here, label precision indicates the ratio of the number of extracted confident examples with correct labels in the total confident example set, and label recall represents the ratio of the number of confident examples with correct labels among the total correctly labeled examples. Specifically, we train a neural network on CIFAR-10  with different kinds and levels of label noise for $25$ epochs respectively and report the performance for each case before and after the proposed PES is applied. 

Results in Table \ref{tab:pre} clearly show that, compared with the traditional early stopping, PES can help to obtain higher accuracies, precisions, and recalls for most cases.  For instance-dependent label noise, PES can achieve higher recall values with comparable label precision values. Note that models with high recall values can help to collect more confident examples, which is critical for learning with confident examples and semi-supervised learning. Therefore, by enhancing the performance of the initial model, PES can help to improve the final classification performance in all cases, which is also verified by the experiments in Section~\ref{sec:3.3}.

\begin{table}[!tp]
\centering
\fontsize{8.5}{10}\selectfont
\caption{Comparison with state-of-the-art methods without semi-supervised learning on CIFAR-10 and CIFAR-100. The mean and standard deviation computed over five runs are presented.}
\vspace{5pt}
\scalebox{1}{{
\begin{tabular}{c | c | c | c | c | c | c}
\Xhline{2\arrayrulewidth}
\multirow{2}{*}{Dataset}  & \multirow{2}{*}{Method}             &  \multicolumn{2}{c |}{Symmetric}       & Pairflip             & \multicolumn{2}{c }{Instance}            \\ \cline{3-7}
	                      &                 &     20\%          &     50\%           &   45\%               &     20\%          &   40\%      \\ \hline
\multirow{7}{*}{CIFAR10} & CE              & 84.00$\pm$0.66    & 75.51$\pm$1.24     & 63.34$\pm$6.03       & 85.10$\pm$0.68    & 77.00$\pm$2.17    \\ 
	                      & Co-teaching     & 87.16$\pm$0.11    & 72.80$\pm$0.45     & 70.11$\pm$1.16       & 86.54$\pm$0.11    & 80.98$\pm$0.39    \\ 
	                      & Forward         & 85.63$\pm$0.52    & 77.92$\pm$0.66     & 60.15$\pm$1.97       & 85.29$\pm$0.38    & 74.72$\pm$3.24    \\ 
	                      & Joint Optim     & 89.70$\pm$0.11    & 85.00$\pm$0.17     & 82.63$\pm$1.38       & 89.69$\pm$0.42    & 82.62$\pm$0.57    \\ 
	                      & T-revision      & 89.63$\pm$0.13    & 83.40$\pm$0.65     & 77.06$\pm$6.47       & 90.46$\pm$0.13    & 85.37$\pm$3.36    \\ 
	                      & DMI             & 88.18$\pm$0.36    & 78.28$\pm$0.48     & 57.60$\pm$14.56      & 89.14$\pm$0.36    & 84.78$\pm$1.97    \\ 
	                      & CDR             & 89.72$\pm$0.38    & 82.64$\pm$0.89     & 73.67$\pm$0.54       & 90.41$\pm$0.34    & 83.07$\pm$1.33    \\ \cline{2-7}
                          & Ours                                & \textbf{92.38$\pm$0.40}  & \textbf{87.45$\pm$0.35}  & \textbf{88.43$\pm$1.08} & \textbf{92.69$\pm$0.44}  & \textbf{89.73$\pm$0.51} \\ 
                          \hline \hline
\multirow{7}{*}{CIFAR100} & CE            & 51.43$\pm$0.58    & 37.69$\pm$3.45     & 34.10$\pm$2.04       & 52.19$\pm$1.42      & 42.26$\pm$1.29    \\ 
	                      & Co-teaching   & 59.28$\pm$0.47    & 41.37$\pm$0.08     & 33.22$\pm$0.48       & 57.24$\pm$0.69      & 45.69$\pm$0.99    \\ 
	                      & Forward       & 57.75$\pm$0.37    & 44.66$\pm$1.01     & 27.88$\pm$0.80       & 58.76$\pm$0.66      & 44.50$\pm$0.72    \\ 
	                      & Joint Optim   & 64.55$\pm$0.38    & 50.22$\pm$0.41     & 42.61$\pm$0.61       & 65.15$\pm$0.31      & 55.57$\pm$0.41    \\    
	                      & T-revision    & 65.40$\pm$1.07    & 50.24$\pm$1.45     & 41.10$\pm$1.95       & 60.71$\pm$0.73      & 51.54$\pm$0.91    \\ 
	                      & DMI           & 58.73$\pm$0.70    & 44.25$\pm$1.14     & 26.90$\pm$0.45       & 58.05$\pm$0.20      & 47.36$\pm$0.68    \\ 
	                      & CDR           & 66.52$\pm$0.24    & 55.30$\pm$0.96     & 43.87$\pm$1.35       & 67.33$\pm$0.67      & 55.94$\pm$0.56    \\ \cline{2-7}
                          & Ours          & \textbf{68.89$\pm$0.45}  & \textbf{58.90$\pm$2.72}  & \textbf{57.18$\pm$1.44}  & \textbf{70.49$\pm$0.79} & \textbf{65.68$\pm$1.41}    \\ 
\Xhline{2\arrayrulewidth}
\end{tabular}
}}
\label{tab:cifar_wo}
\vspace{-5pt}
\end{table}

\begin{table}[!tp]
\fontsize{8.5}{10}\selectfont
\centering
\caption{Comparison with state-of-the-art methods with semi-supervised learning on CIFAR-10 and CIFAR-100 with symmetric label noise from different levels. Results with * are token from \cite{Li2020DivideMix}. The mean and standard deviation are computed over three runs.}
\vspace{5pt}
\scalebox{1}{
{
\begin{tabular}{c | c | c | c | c | c | c }
\Xhline{2\arrayrulewidth}
	Dataset             &  \multicolumn{3}{c |}{CIFAR-10}                  & \multicolumn{3}{c}{CIFAR-100}                              \\ \hline
	Methods / Noise     & Sym-20\%       & Sym-50\%       & Sym-80\%       & Sym-20\%           & Sym-50\%          & Sym-80\%          \\ \hline
	% CE                  & 87.2           & 80.9           & 65.8           & 58.1               & 47.5              & 23.6              \\
	CE					& 86.5$\pm$0.6   & 80.6$\pm$0.2   & 63.7$\pm$0.8   & 57.9$\pm$0.4       & 47.3$\pm$0.2      & 22.3$\pm$1.2      \\ 
	% MixUp               & 93.5           & 88.4           & 73.6           & 69.7               & 57.9              & 34.69             \\
	MixUp				    & 93.2$\pm$0.3   & 88.2$\pm$0.3   & 73.3$\pm$0.3   & 69.5$\pm$0.2       & 57.1$\pm$0.6      & 34.1$\pm$0.6      \\ 
	M-correction*       & 94.0           & 92.0           & 86.8           & 73.9               & 66.1              & 48.2              \\ 
	DivideMix*          & 95.2           & 94.2           & 93.0           & 75.2               & 72.8              & 58.3              \\ 
	% DivideMix (ensemble)*\cite{Li2020DivideMix} & \textbf{96.1} & 94.6  & 93.2           & 77.3               & 74.6              & 60.2              \\ \hline
	% DivideMix           & 95.8           & 94.6           & 93.1           & 75.3               & 73.1              & 56.9              \\ 
	DivideMix           & 95.6$\pm$0.1   & 94.6$\pm$0.1   & 92.9$\pm$0.3   & 75.3$\pm$0.1       & 72.7$\pm$0.6      & 56.4$\pm$0.3      \\ 
	% ELR+              & 95.0           & 93.6           & 90.5           & 75.7               & 71.1              & 51.28             \\ 
	ELR+                    & 94.9$\pm$0.2   & 93.6$\pm$0.1   & 90.4$\pm$0.2   & 75.5$\pm$0.2       & 71.0$\pm$0.2      & 50.4$\pm$0.8      \\ \hline
    % Ours (Semi)         & \textbf{96.1}  & \textbf{95.3}  & \textbf{93.3}  & \textbf{77.7}      & \textbf{74.9}     & \textbf{62.3}     \\
    Ours (Semi)         & \textbf{95.9$\pm$0.1}  & \textbf{95.1$\pm$0.2}   & \textbf{93.1$\pm$0.2} & \textbf{77.4$\pm$0.3} & \textbf{74.3$\pm$0.6}  & \textbf{61.6$\pm$0.6}      \\ 
\Xhline{2\arrayrulewidth}
\end{tabular}
}}
\label{tab:cifar_sym}
\vspace{-10pt}
\end{table}

\subsection{Classification Accuracy Evaluation}
\label{sec:3.3}
\textbf{Synthetic datasets.}  We first verify the effectiveness of our proposed method without semi-supervised learning techniques on two synthetic datasets: CIFAR-10 and CIFAR-100. For both of these two datasets, we leave 10\% of data with noisy labels as noisy validation set. Results are presented in Table \ref{tab:cifar_wo}, which shows that our proposed  method can consistently outperform all other baselines across various settings by a large margin.

\begin{table}[!tp]
\fontsize{8.5}{10}\selectfont
\centering
\caption{Comparison with state-of-the-art methods with semi-supervised learning on CIFAR-10 and CIFAR-100 with instance-dependent and pairflip label noise from different levels. The mean and standard deviation are computed over three runs.}
\vspace{5pt}
\scalebox{1}{
\begin{tabular}{c | c | c | c | c | c| c} 
\Xhline{2\arrayrulewidth}
	Dataset          &  \multicolumn{3}{c |}{CIFAR-10}                  & \multicolumn{3}{c}{CIFAR-100}                              \\ \hline
	Methods / Noise  & Inst-20\%       & Inst-40\%      & Pair-45\%    & Inst-20\%        & Inst-40\%      & Pair-45\%       \\ \hline
	CE               & 87.5$\pm$0.5    & 78.9$\pm$0.7   & 74.9$\pm$1.7 & 56.8$\pm$0.4     & 48.2$\pm$0.5   & 38.5$\pm$0.6    \\
	MixUp            & 93.3$\pm$0.2    & 87.6$\pm$0.5   & 82.4$\pm$1.0 & 67.1$\pm$0.1     & 55.0$\pm$0.1   & 44.2$\pm$0.5    \\
	DivideMix        & 95.5$\pm$0.1    & 94.5$\pm$0.2   & 85.6$\pm$1.7 & 75.2$\pm$0.2     & 70.9$\pm$0.1   & 48.2$\pm$1.0    \\
	ELR+             & 94.9$\pm$0.1    & 94.3$\pm$0.2   & 86.1$\pm$1.2 & 75.8$\pm$0.1     & 74.3$\pm$0.3   & 65.3$\pm$1.3    \\ \hline
	Ours (Semi)      & \textbf{95.9$\pm$0.1}  & \textbf{95.3$\pm$0.1}   & \textbf{94.5$\pm$0.3}  & \textbf{77.6$\pm$0.3} & \textbf{76.1$\pm$0.4} & \textbf{73.6$\pm$1.7} \\
\Xhline{2\arrayrulewidth}
\end{tabular}
}
\label{tab:cifar_asym}
\end{table}

Table \ref{tab:cifar_sym} and Table \ref{tab:cifar_asym} present the mean accuracy and standard deviation for our method and all baselines on CIFAR-10 and CIFAR-100, respectively. From the results, we can get that the proposed method can outperform all baselines in all cases. For pairflip label noise, the advantages of our proposed method become more apparent, and it significantly outperforms state-of-the-art methods by over 8\% on both CIFAR-10 and CIFAR-100. These empirical results support our proposal that PES can improve the quality of selected confident examples, which helps improve performance and reduce the variance of the final classifier. 

\textbf{Real-world dataset.} We evaluate the performance of the proposed method on a real-world dataset with Clothing-1M~\cite{Xiao2015Clothing} and select methods such as CE, Forward, Joint-Optim, DMI, and T-revision, which use a single network, and also methods such as DivideMix and ELR+, which adopt an ensemble model with two different networks, as baselines. We also report the results for the proposed PES with a single network as \emph{ours} and the results for PES, which ensembles two networks, as \emph{ours*}. The overall results are reported in Table \ref{tab:clothing}, from which we can observe that the proposed PES with a single network can outperform all baselines using a single network. And with an ensemble model, which contains two different networks, our method can outperform all the adopted baselines. These results clearly demonstrate that, by improving the performance of the initial classification network, our method is more flexible to handle such real-world noise problems.

\begin{table}[!tp]
\centering
\fontsize{8.5}{10}\selectfont
\label{tab:cifar100}
\caption{Compassion with state-of-the-art methods on Clothing-1M. Results of baseline methods are taken from the original papers. ours represent the results obtained by PES with a single network and ours* indicate the results obtained by PES with an ensemble model. }
\vspace{5pt}
\scalebox{1}{
{
\begin{tabular}{ c | c | c | c | c | c | c | c | c}
\Xhline{2\arrayrulewidth}
    CE       & Forward  &  Joint-Optim  &   DMI    &  T-revision & DivideMix* & ELR+*  &  Ours          &  Ours*         \\ \hline 
    69.21    & 69.84    &  72.16        & 72.46    &  74.18      & 74.76      & 74.81  & \textbf{74.64} & \textbf{74.99} \\
\Xhline{2\arrayrulewidth}
\end{tabular} 
}}
\label{tab:clothing}
\end{table}

\begin{figure}[!t]
\centering
\subfloat[$T_2$\label{fig:ablation}]{\includegraphics[width=0.33\textwidth]{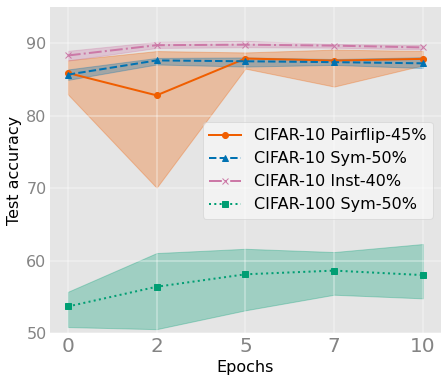}} 
\subfloat[$T_3$\label{fig:ablation2}]{\includegraphics[width=0.33\textwidth]{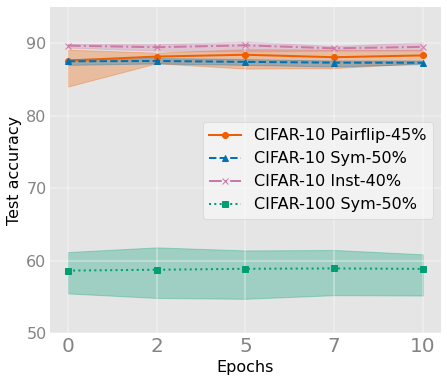}} 
\caption{Sensitivity analysis for different training iteration numbers: $T_2$ and $T_3$.} 
\vspace{-10pt}
\end{figure}

\subsection{Sensitivity Analysis}
\label{sec:abl}
In this section, we investigate the hyper-parameter sensitivity for the training iteration number $T_2$ and $T_3$, respectively. We firstly analyze the training epoch number for the second DNN part by varying $T_2$ from the range of $[0,10]$. The results are illustrated in Figure \ref{fig:ablation}, from which can find that, with the increasing of $T_2$, the performance of PES first increase and then decrease in all the cases except for 45\% Pairflip noise on the CIFAR-10 dataset. While the model achieves the best performance with $T_2$ as $7$ for all types of noisy labels.  Then we fix $T_2$ as 7, and analyze the impact of  the third DNN part by varying $T_3$ from the range of $[0,10]$. The results are shown in Figure \ref{fig:ablation2}. Although the performance variance for different $T_3$ is smaller than that for $T_2$, we can still observe that the best performance can be obtained when $T_3$ is set as $5$. More importantly, from these two figures, we can get that both $T_2$ and $T_3$ are robust to the different types of noisy labels.

\subsection{Training Time Comparison}
\label{subsec:time}
In this section, we compare the training time of our method and other state-of-the-art baselines. All the experiments are conducted on a server with a single Nvidia V100 GPU. The training times for all the methods are reported in Table \ref{tab:training_time}, from which we can get that our algorithm with cross-entropy loss achieves the fastest speed across all baselines, only about 1 hour. Our method combining with MixMatch \cite{Berthelot2019MixMatch} is also fast, only a little more than half of the training time of DivideMix. The time of ELR+ \cite{Liu2020ELR} shows superior, but ELR+ trains the network with fewer epochs, with 200 epochs compared with ours for 300 epochs. 

\begin{table}[!tp]
\fontsize{8.5}{10}\selectfont
\centering
\caption{Training time comparison for baselines on CIFAR-10 with 50\% Symmetric label noise.}
\label{tab:training_time}
\vspace{5pt}
\scalebox{1}{{
\begin{tabular}{c | c | c | c | c | c | c | c}
\Xhline{2\arrayrulewidth}
	CE    & Co-teaching   & CDR         & T-revision  & ELR+     & DivideMix & Ours  & Ours (Semi)  \\ \hline
	0.9h  & 1.5h          & 3.0h        & 3.5h        & 2.2h     & 5.5h      & 1.0h  & 3.1h         \\
\Xhline{2\arrayrulewidth}
\end{tabular} 
}}
\vspace{-10pt}
\end{table}

\section{Related work}
\label{sec:rel}
Learning with noisy data has been well studied \cite{liu2016reweighting, Goldberger17AdaptLayer, Ma18Dimensionality, Wang2019Symmetric, Ma2020NormalizedLoss}. Current works can be mainly categorized into two groups: model-based and model-free methods. In this section, we briefly review some closely related works. 

The first type models the relationship between clean labels and noisy labels by estimating the noise transition matrix and build a loss function to correct the loss \cite{Patrini2017forward, xia2020part, Yao2020Dual, wu2020class2simi}. \cite{Patrini2017forward} first combines algorithms for estimating the noise rates and loss correction techniques together and introduces two alternative procedures for loss correction. It also proves that both of the two procedures enjoy formal robustness guarantees \emph{w.r.t.} the clean data distribution. DMI \cite{Xu2019DMI} proposes an information-theoretic loss function, which utilizes Shannon’s mutual information and is robustness to different kinds of label noise. T-revision \cite{xia2019anchor} estimates the noise transition matrix without anchor points by adding a fine-tuned slack variables. Although these methods have made certain progress, they are usually fragile to estimate the noise transition matrix for heavy noisy data and are also hard to handle a large number of classes. Therefore, in this paper, we mainly focus on the model-free methods.

The second strand mainly counteracts noisy labels by exploiting the memorization effect that deep networks tend to first memorize and fit majority (clean) patterns and then overfit minority (noisy) patterns~\cite{Arpit2017Look}. To exploit this property,  Co-teaching \cite{han2018co} employs two networks with different initialization and uses \textit{small loss} to select confident examples.  M-correction \cite{Arazo2019UnsupervisedLabel} uses two Gaussian Mixture Models to identify confident examples, instead of using networks themselves. DivideMix \cite{Li2020DivideMix} extends Co-teaching \cite{han2018co} and employs two Beta Mixture Model to select confident examples. MixMatch \cite{Berthelot2019MixMatch} is then adopted to leverage unconfident examples with a semi-supervised learning framework. All the above methods exploit the memorization effect by considering the adopted network as a whole. Recently, \cite{Jimmy2020GoodRepr} shows that networks training with noisy labels can produce good representations, if the structure of networks suits the targeted tasks. Our method further explains that noisy labels have different impacts for different layers in a DNN. And latter layers will receive earlier and more severe impact than their former counterparts. Therefore, by considering a DNN as a composition of several layers and training different layers with different epochs, our method is able to better exploit the memorization effect and achieve superior performance.

%%%%------------------------------------------------------
\section{Conclusion}
\label{sec:con}
In this work, we provide a progressive early stopping (PES) method to better exploit the memorization effect of deep neural networks (DNN) for noisy-label learning. We first find that the impact of noisy labels for former layers in a DNN is much less and later than that for latter DNN layers, and then build upon this insight to propose the PES method, which separates a DNN into different parts and progressively train each part to counteract the different impacts of noisy labels for different DNN layers. To show that PES can boost the performance of state-of-the-art methods, we conduct extensive experiments across multiple synthetic and real-world noisy datasets and demonstrate that the proposed PES can help to obtain substantial performance improvements compared to current state-of-the-art baselines. 
The main limitation of our method lies in that, by splitting a DNN into different parts, PES introduces several additional hyper-parameters that need to be tuned carefully. 
In the future, we will extend the work in the following aspects. First, we will study other mechanisms that distinguishing desired and undesired memorization rather than early stopping, e.g., the gradient ascent trick \cite{han2020SIGUA}. Second, we are interested in combining PES with interesting ideas from semi-supervised learning and unsupervised learning.

%%%%------------------------

\begin{ack}
YB was partially supported by Agriculture Consultant and Smart Management. BH was supported by the RGC Early Career Scheme No. 22200720, NSFC Young Scientists Fund No. 62006202 and HKBU CSD Departmental Incentive Grant. YY was partially supported by Key Research and Development Program of Shaanxi (ProgramNo. 2021ZDLGY01-03). GN was supported by JST AIP Acceleration Research Grant Number JPMJCR20U3, Japan. TL was partially supported by Australian Research Council Projects DE-190101473 and IC-190100031.

\end{ack}

\clearpage
\newpage
{\small
\bibliographystyle{plain}
\bibliography{egbib}

\begin{thebibliography}{10}

\bibitem{Arazo2019UnsupervisedLabel}
Eric Arazo, Diego Ortego, Paul Albert, Noel~E. O'Connor, and Kevin McGuinness.
\newblock Unsupervised label noise modeling and loss correction.
\newblock In {\em ICML}, pages 312--321, 2019.

\bibitem{Arpit2017Look}
Devansh Arpit, Stanislaw Jastrzebski, Nicolas Ballas, David Krueger, Emmanuel
  Bengio, Maxinder~S. Kanwal, Tegan Maharaj, Asja Fischer, Aaron~C. Courville,
  Yoshua Bengio, and Simon Lacoste{-}Julien.
\newblock A closer look at memorization in deep networks.
\newblock In {\em ICML}, pages 233--242, 2017.

\bibitem{belkin2019reconciling}
Mikhail Belkin, Daniel Hsu, Siyuan Ma, and Soumik Mandal.
\newblock Reconciling modern machine-learning practice and the classical
  bias--variance trade-off.
\newblock {\em Proceedings of the National Academy of Sciences},
  116(32):15849--15854, 2019.

\bibitem{Berthelot2019MixMatch}
David Berthelot, Nicholas Carlini, Ian~J. Goodfellow, Nicolas Papernot, Avital
  Oliver, and Colin Raffel.
\newblock Mixmatch: {A} holistic approach to semi-supervised learning.
\newblock In {\em NeurIPS}, pages 5050--5060, 2019.

\bibitem{cha2012social}
Youngchul Cha and Junghoo Cho.
\newblock Social-network analysis using topic models.
\newblock In {\em SIGIR}, pages 565--574, 2012.

\bibitem{Cun1989Handwritten}
Y.~Le Cun, Larry~D. Jackel, Bernhard~E. Boser, John~S. Denker, Hans~Peter Graf,
  Isabelle Guyon, Donnie Henderson, Richard~E. Howard, and Wayne~E. Hubbard.
\newblock Handwritten digit recognition: Applications of neural net chips and
  automatic learning.
\newblock In {\em {NATO}}, pages 303--318, 1989.

\bibitem{Goldberger17AdaptLayer}
Jacob Goldberger and Ehud Ben{-}Reuven.
\newblock Training deep neural-networks using a noise adaptation layer.
\newblock In {\em {ICLR}}, 2017.

\bibitem{goodfellow2016deep}
Ian Goodfellow, Yoshua Bengio, Aaron Courville, and Yoshua Bengio.
\newblock {\em Deep learning}.
\newblock The MIT Press, 2016.

\bibitem{han2020SIGUA}
Bo~Han, Gang Niu, Xingrui Yu, Quanming Yao, Miao Xu, Ivor~W. Tsang, and Masashi
  Sugiyama.
\newblock {SIGUA:} forgetting may make learning with noisy labels more robust.
\newblock In {\em ICML}, pages 4006--4016, 2020.

\bibitem{han2018co}
Bo~Han, Quanming Yao, Xingrui Yu, Gang Niu, Miao Xu, Weihua Hu, Ivor Tsang, and
  Masashi Sugiyama.
\newblock Co-teaching: Robust training of deep neural networks with extremely
  noisy labels.
\newblock In {\em NeurIPS}, pages 8527--8537, 2018.

\bibitem{He2016ResNet}
Kaiming He, Xiangyu Zhang, Shaoqing Ren, and Jian Sun.
\newblock Deep residual learning for image recognition.
\newblock In {\em CVPR}, pages 770--778, 2016.

\bibitem{ishida2020we}
Takashi Ishida, Ikko Yamane, Tomoya Sakai, Gang Niu, and Masashi Sugiyama.
\newblock Do we need zero training loss after achieving zero training error?
\newblock In {\em ICML}, pages 4604--4614, 2020.

\bibitem{krizhevsky2009CIFAR}
Alex Krizhevsky, Geoffrey Hinton, et~al.
\newblock Learning multiple layers of features from tiny images.
\newblock Technical report, 2009.

\bibitem{krizhevsky2012imagenet}
Alex Krizhevsky, Ilya Sutskever, and Geoffrey~E Hinton.
\newblock Imagenet classification with deep convolutional neural networks.
\newblock In {\em NeurIPS}, pages 1097--1105, 2012.

\bibitem{Jimmy2020GoodRepr}
Jingling Li, Mozhi Zhang, Keyulu Xu, John~P. Dickerson, and Jimmy Ba.
\newblock Noisy labels can induce good representations.
\newblock {\em arXiv preprint arXiv:2012.12896}, 2020.

\bibitem{Li2020DivideMix}
Junnan Li, Richard Socher, and Steven C.~H. Hoi.
\newblock Dividemix: Learning with noisy labels as semi-supervised learning.
\newblock In {\em ICLR}, 2020.

\bibitem{Liu2020ELR}
Sheng Liu, Jonathan Niles{-}Weed, Narges Razavian, and Carlos
  Fernandez{-}Granda.
\newblock Early-learning regularization prevents memorization of noisy labels.
\newblock In {\em NeurIPS}, pages 20331--20342, 2020.

\bibitem{liu2016reweighting}
Tongliang Liu and Dacheng Tao.
\newblock Classification with noisy labels by importance reweighting.
\newblock {\em IEEE Transactions on pattern analysis and machine intelligence},
  38(3):447--461, 2016.

\bibitem{liu2011noise}
Wei Liu, Yu-Gang Jiang, Jiebo Luo, and Shih-Fu Chang.
\newblock Noise resistant graph ranking for improved web image search.
\newblock In {\em CVPR}, pages 849--856, 2011.

\bibitem{Loshchilov2017CosineLr}
Ilya Loshchilov and Frank Hutter.
\newblock {SGDR:} stochastic gradient descent with warm restarts.
\newblock In {\em ICLR}, 2017.

\bibitem{Lyu20CurriculumLoss}
Yueming Lyu and Ivor~W. Tsang.
\newblock Curriculum loss: Robust learning and generalization against label
  corruption.
\newblock In {\em {ICLR}}, 2020.

\bibitem{Ma2020NormalizedLoss}
Xingjun Ma, Hanxun Huang, Yisen Wang, Simone Romano, Sarah~M. Erfani, and James
  Bailey.
\newblock Normalized loss functions for deep learning with noisy labels.
\newblock In {\em {ICML}}, pages 6543--6553, 2020.

\bibitem{Ma18Dimensionality}
Xingjun Ma, Yisen Wang, Michael~E. Houle, Shuo Zhou, Sarah~M. Erfani, Shu{-}Tao
  Xia, Sudanthi N.~R. Wijewickrema, and James Bailey.
\newblock Dimensionality-driven learning with noisy labels.
\newblock In {\em {ICML}}, pages 3361--3370, 2018.

\bibitem{nakkiran2020deep}
Preetum Nakkiran, Gal Kaplun, Yamini Bansal, Tristan Yang, Boaz Barak, and Ilya
  Sutskever.
\newblock Deep double descent: Where bigger models and more data hurt.
\newblock In {\em ICLR}, 2020.

\bibitem{Nguyen2020SELF}
Duc~Tam Nguyen, Chaithanya~Kumar Mummadi, Thi{-}Phuong{-}Nhung Ngo, Thi
  Hoai~Phuong Nguyen, Laura Beggel, and Thomas Brox.
\newblock {SELF:} learning to filter noisy labels with self-ensembling.
\newblock In {\em ICLR}, 2020.

\bibitem{Patrini2017forward}
Giorgio Patrini, Alessandro Rozza, Aditya~Krishna Menon, Richard Nock, and
  Lizhen Qu.
\newblock Making deep neural networks robust to label noise: {A} loss
  correction approach.
\newblock In {\em CVPR}, pages 2233--2241, 2017.

\bibitem{tanaka2018joint}
Daiki Tanaka, Daiki Ikami, Toshihiko Yamasaki, and Kiyoharu Aizawa.
\newblock Joint optimization framework for learning with noisy labels.
\newblock In {\em CVPR}, pages 5552--5560, 2018.

\bibitem{Wang2018Iterative}
Yisen Wang, Weiyang Liu, Xingjun Ma, James Bailey, Hongyuan Zha, Le~Song, and
  Shu{-}Tao Xia.
\newblock Iterative learning with open-set noisy labels.
\newblock In {\em {CVPR}}, pages 8688--8696, 2018.

\bibitem{Wang2019Symmetric}
Yisen Wang, Xingjun Ma, Zaiyi Chen, Yuan Luo, Jinfeng Yi, and James Bailey.
\newblock Symmetric cross entropy for robust learning with noisy labels.
\newblock In {\em {ICCV}}, pages 322--330, 2019.

\bibitem{wu2020class2simi}
Songhua Wu, Xiaobo Xia, Tongliang Liu, Bo~Han, Mingming Gong, Nannan Wang,
  Haifeng Liu, and Gang Niu.
\newblock Class2simi: A noise reduction perspective on learning with noisy
  labels.
\newblock In {\em ICML}, pages 11285--11295, 2021.

\bibitem{xia2021robust}
Xiaobo Xia, Tongliang Liu, Bo~Han, Chen Gong, Nannan Wang, Zongyuan Ge, and
  Yi~Chang.
\newblock Robust early-learning: Hindering the memorization of noisy labels.
\newblock In {\em ICLR}, 2021.

\bibitem{xia2020part}
Xiaobo Xia, Tongliang Liu, Bo~Han, Nannan Wang, Mingming Gong, Haifeng Liu,
  Gang Niu, Dacheng Tao, and Masashi Sugiyama.
\newblock Part-dependent label noise: Towards instance-dependent label noise.
\newblock In {\em NeurIPS}, pages 7597--7610, 2020.

\bibitem{xia2019anchor}
Xiaobo Xia, Tongliang Liu, Nannan Wang, Bo~Han, Chen Gong, Gang Niu, and
  Masashi Sugiyama.
\newblock Are anchor points really indispensable in label-noise learning?
\newblock In {\em NeurIPS}, pages 6835--6846, 2019.

\bibitem{Xiao2017Fashion}
Han Xiao, Kashif Rasul, and Roland Vollgraf.
\newblock Fashion-mnist: a novel image dataset for benchmarking machine
  learning algorithms.
\newblock {\em arXiv preprint arXiv:1708.07747}, 2017.

\bibitem{Xiao2015Clothing}
Tong Xiao, Tian Xia, Yi~Yang, Chang Huang, and Xiaogang Wang.
\newblock Learning from massive noisy labeled data for image classification.
\newblock In {\em CVPR}, pages 2691--2699, 2015.

\bibitem{Xie2020UDA}
Qizhe Xie, Zihang Dai, Eduard~H. Hovy, Thang Luong, and Quoc Le.
\newblock Unsupervised data augmentation for consistency training.
\newblock In {\em NeurIPS}, pages 6256--6268, 2020.

\bibitem{Xu2019DMI}
Yilun Xu, Peng Cao, Yuqing Kong, and Yizhou Wang.
\newblock L{\_}dmi: {A} novel information-theoretic loss function for training
  deep nets robust to label noise.
\newblock In {\em NeurIPS}, pages 6222--6233, 2019.

\bibitem{Yao2020Dual}
Yu~Yao, Tongliang Liu, Bo~Han, Mingming Gong, Jiankang Deng, Gang Niu, and
  Masashi Sugiyama.
\newblock Dual {T:} reducing estimation error for transition matrix in
  label-noise learning.
\newblock In {\em NeurIPS}, pages 7260--7271, 2020.

\bibitem{Yu19Disagreement}
Xingrui Yu, Bo~Han, Jiangchao Yao, Gang Niu, Ivor~W. Tsang, and Masashi
  Sugiyama.
\newblock How does disagreement help generalization against label corruption?
\newblock In {\em ICML}, pages 7164--7173, 2019.

\bibitem{zhang2016understanding}
Chiyuan Zhang, Samy Bengio, Moritz Hardt, Benjamin Recht, and Oriol Vinyals.
\newblock Understanding deep learning requires rethinking generalization.
\newblock In {\em ICLR}, 2016.

\bibitem{Zhang2017MixUp}
Hongyi Zhang, Moustapha Ciss{\'{e}}, Yann~N. Dauphin, and David Lopez{-}Paz.
\newblock mixup: Beyond empirical risk minimization.
\newblock In {\em {ICLR}}, 2018.

\end{thebibliography}
}

\clearpage
\newpage

\appendix
\section{Training details}
In this section, we first provide details about the adopted three kinds of noisy labels. Then, we elaborate on the data preprocessing and the hyperparameter settings in our experiments.

\subsection{Definition of noise}
According to different correlations between noisy labels and clean labels, there are three kinds of widely used label noise, namely symmetric class-dependent label noise, pairflip class-dependent label noise, and instance-dependent label noise~\cite{Patrini2017forward, han2018co, xia2020part}. In the following, we first introduce one basic concept: transition matrix~\cite{Patrini2017forward}, and then provide the details for all the three kinds of label noise, respectively. %synthetic label noise used in the paper, which have been widely studied in many works \cite{Patrini2017forward, han2018co, Yao2020Dual, xia2020parts}. The generation methods are as follows.

\textbf{Transition matrix:}\ \  The \textit{transition matrix} $T(\bm{x})$ is used to explicitly model the generation process of label noise, where $T_{ij}(\bm{x})=\text{Pr}(\bar{Y}=j|Y=i,X=\bm{x})$ is the flip rate between the true label  and noisy label on given
data $\bm{x}$. $X$ is the variable of instances, ${Y}$ is the variable of clean labels, and $\bar{Y}$ is the variable of noisy labels. $T_{ij}(\bm{x})$ is the $ij$-th entry of the transition matrix $T(\bm{x})$, which denotes the probability of the instance $\bm{x}$ with clean label $i$ being observed with a noisy label $j$.%  will flip from the clean class $j$ to the noisy class $i$.

\textbf{Symmetric class-dependent label noise:}\ \ Symmetric class-dependent label noise is generated with symmetric class-dependent noise transition matrices. We set the flip rate $\alpha$. Random flipping labels may change to true labels, so the flip rate may include or exclude true labels. For the flip rate excluding true labels, the diagonal entries of symmetric transition matrix are $1-\alpha$ and the off-diagonal entries are $\alpha/(c-1)$. For the flip rate including true labels, the diagonal entries of symmetric transition matrix are $1-(\alpha \times (c-1)/c)$ and the off-diagonal entries are $\alpha/c$.

\textbf{Pairflip class-dependent label noise:}\ \ Pairflip noise is a simulation of fine-grained classification with noisy labels, where annotators may make mistakes only within very similar classes\cite{Yu19Disagreement, Lyu20CurriculumLoss}. The label noise is generated with pairflip class-dependent noise transition matrices, which is defined as follow. Let flip rate is $\alpha$. The diagonal entries of a pairflip transition matrix are $1-\alpha$ and the entities for their adjacent classes, which the examples in a given class may be wrongly classified to, are $\alpha$.

\textbf{Instance-dependent label noise:} We generate the instance-dependent label noise according to  Algorithm \ref{alg:noise}. More details about this algorithm can be found in \cite{xia2020part}.

\begin{algorithm}[h!]
 {\bfseries Input}: Clean samples $\{(\bm{x}_i, y_i)\}_{i=1}^{n}$; Noise rate $\tau$.
 
	1: Sample instance flip rates $q\in\mathbb{R}^{n}$ from the truncated normal distribution $\tau\mathcal{N}(\tau,0.1^2,[0,1])$;
	
	2: Independently sample $w_1,w_2,\ldots,w_c$ from the standard normal distribution $\mathcal{N}(0,1^2)$;
	
	3: For $i=1,2,\ldots,n$ do
	
    4:\quad $p=\bm{x}_i\times w_{y_i}$; \hfill// generate instance-dependent flip rates
    
    5:\quad  $p_{y_i}=-\infty$; \hfill// control the diagonal entry of the instance-dependent transition matrix
    
    6:\quad  $p=q_i\times softmax(p)$; \hfill// make the sum of the off-diagonal entries of the $y_i$-th row to be $q_i$
    
    7:\quad  $p_{y_i}=1-q_i$; \hfill// set the diagonal entry to be $1-q_i$
    
    8:\quad  Randomly choose a label from the label space according to possibilities $p$ as noisy label $\bar{y}_i$;
    
	9: End for.
	
{\bfseries Output}: Noisy samples $\{(\bm{x}_i, \bar{y}_i)\}_{i=1}^{n}$
\caption{Instance-dependent Label Noise Generation}
\label{alg:noise}
\end{algorithm}

\subsection{Data preprocessing and experimental settings}
\textbf{Data preprocessing:} For experiments on CIFAR-10/100 \cite{krizhevsky2009CIFAR} without semi-supervised learning, we use simple data augmentation techniques including random crop and horizontal flip. For experiments on CIFAR-10/100 with semi-supervised learning, except random cropping and horizontal flip, MixUp \cite{Zhang2017MixUp} is also employed, which is a critical component of MixMatch \cite{Berthelot2019MixMatch}. For Clothing-1M \cite{Xiao2015Clothing}, we first resize images to 256 × 256, and then random crop to 224 × 224, following a random horizontal flip. 

\textbf{Hyper-parameters of PES}:  We adopt an Adam optimizer for $T_2$ and $T_3$ for accelerating the model training and reducing the parameter turning, and $T_2$ and $T_3$ are chosen from \{2, 5, 7\}. Note that the number of total training epochs includes $T_1$, but excludes $T_2$ and $T_3$. To make PES work in large datasets, we regard training $100,000$ examples as an epoch in Clothing1M experiments. 

\textbf{Hyper-parameters of semi-supervised learning}: We keep all the hyper-parameters fixed for different levels of noise, and only adjust $\lambda_u$ for different noisy settings, since the ratio of confident examples (labeled data) and unconfident examples (unlabeled data) can vary greatly for different noisy settings. Specifically, we set $ K = 2$, $T = 0.5$, and $\lambda_u$ is chosen from $\{5, 15, 25, 50, 75, 100\}$. $\alpha$ begins with $4$, and changes to $0.75$ after 150th epoch. More details of hyper-parameters can be found in Table \ref{tab:hyper} and Table \ref{tab:lambda_u}. 

\begin{table}[!h]
\vspace{-10pt}
\centering
\caption{Training hyper-parameters for CIFAR-10/100 and Clothing-1M}
\label{tab:hyper}
\vspace{5pt}
\scalebox{0.8}{
\begin{tabular}{c | c | c | c | c | c }
\Xhline{2\arrayrulewidth}
	                   & \multicolumn{2}{c |}{CIFAR-10} & \multicolumn{2}{c |}{CIFAR-100}        & Clothing-1M           \\ \hline
	architecture       & ResNet-18      & PreAct ResNet-18  & ResNet-34      & PreAct ResNet-18  & Pretrained Resnet-50  \\ 
	loss function      & CE             & MixMatch loss     & CE             & MixMatch loss     & CE                    \\   
	learning rate (lr) & 0.1            & 0.02              & 0.1            & 0.02              & $5 \times 10^{-3}$    \\
	lr decay           & 100th \& 150th & Cosine Annealing  & 100th \& 150th & Cosine Annealing  & 20th \& 30th          \\ 
	weight decay       & $10^{-4}$      & $5 \times 10^{-4}$& $10^{-4}$      & $5 \times 10^{-4}$& $10^{-3}$             \\ 
	batch size         & 128            & 128               & 128            & 128               & 64                    \\ 
	training examples  & 45,000         & 50,000            & 45,000         & 50,000            & 1,000,000             \\ 
	training epochs    & 200            & 300               & 200            & 300               & 50                     \\ 
	PES lr             & $10^{-4}$      & $10^{-4}$         & $10^{-4}$      & $10^{-4}$         & $5 \times 10^{-6}$    \\ 
	$T_1$              & 25             & 20                & 30             & 35                & 20                     \\ 
	$T_2$              & 7              & 5                 & 7              & 5                 & 7                     \\ 
	$T_3$              & 5              & -                 & 5              & -                 & -                     \\ 

\Xhline{2\arrayrulewidth}
\end{tabular} 
}
\end{table}

\begin{table}[!h]
\centering
\caption{Semi-supervised loss weight $\lambda_u$ for CIFAR-10/100}
\label{tab:lambda_u}
\vspace{5pt}
\scalebox{0.8}{
\begin{tabular}{c | c | c | c | c | c | c }
\Xhline{2\arrayrulewidth}
	Datasets / Noise    & Sym-20\%  & Sym-50\%  & Sym-80\%  & Pairflip-45\%  & Inst-20\%  & Inst-40\%  \\ \hline
	CIFAR-10            & 5         & 15        & 25        & 5              & 5          & 15         \\ \hline
	CIFAR-100           & 50        & 75        & 100       & 50             & 50         & 50         \\
\Xhline{2\arrayrulewidth}
\end{tabular} 
}
\end{table}

\section{Additional experiments}
In this section, we provide more experimental results on CIFAR-100 and Fashion-MNIST to further verify the hypothesize that noisy labels may have more severe impacts on the latter layers. We also provide additional comparisons with baselines, which exploit ensemble networks.

In the first experiment, we adopt a dataset with more classes: CIFAR-100 and a deeper network: ResNet-34 \cite{He2016ResNet}. In addition, we adopt Fashion-MNIST \cite{Xiao2017Fashion}, including 60,000 training images with 28x28 size and LeNet \cite{Cun1989Handwritten}, which consists of two convolutional layers and three full-connected layers with ReLU activation. The learning procedure for CIFAR-100 and Fashion-MNIST is the same as that for CIFAR-10 in the paper. Specifically, we first train the whole network on noisy data with different training epochs. For the final layer, we directly report the overall classification performance. For other selected layers, we frozen the parameters for the selected layer and previous layers, and then reinitialize and optimize the rest layers with clean data, and the final classification performance is adopted to evaluate the impact of noisy labels. We do not use image augmentation techniques for Fashion-MNIST dataset.

\begin{figure}[!h]
\vspace{-10pt}
\centering
\subfloat[Symmetric 50\%]{{\includegraphics[width=0.33\textwidth]{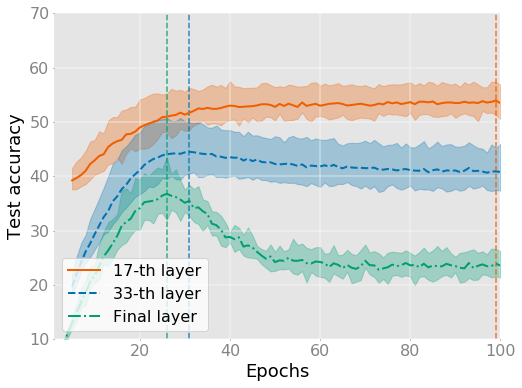}}}
\subfloat[Pairflip 45\%]{{\includegraphics[width=0.33\textwidth]{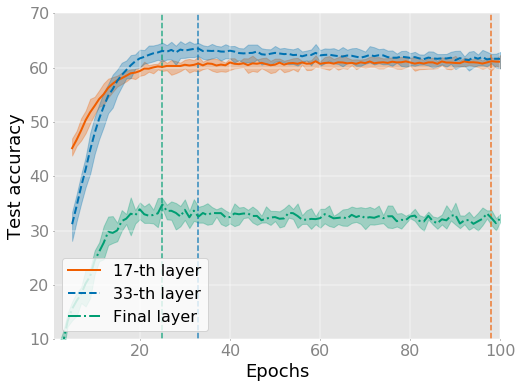}}}
\subfloat[Instance 40\%]{{\includegraphics[width=0.33\textwidth]{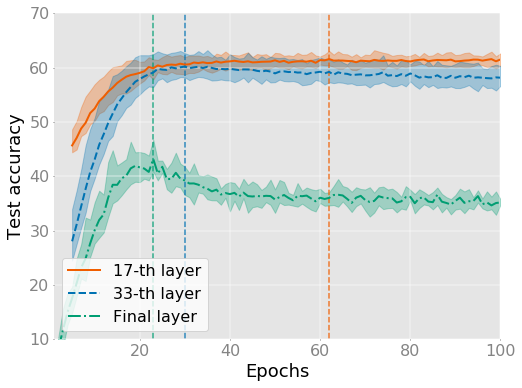}}}
\caption{We adopt ResNet-34 as the model on CIFAR-100 and evaluate the impact of noisy labels on the representations from the $17$-th layer, the $33$-th layer, and the final layer. The curves present the mean of five runs and the
best performances highlight with dotted vertical lines.}
\label{fig:4}
\vspace{-10pt}
\end{figure}

\begin{figure}[!h]
\centering
\subfloat[Symmetric 50\%]{{\includegraphics[width=0.33\textwidth]{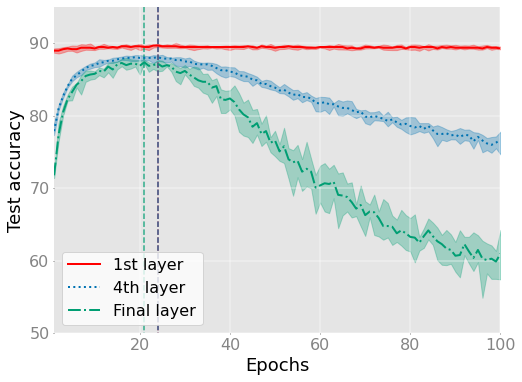}}}
\subfloat[Pairflip 45\%]{{\includegraphics[width=0.33\textwidth]{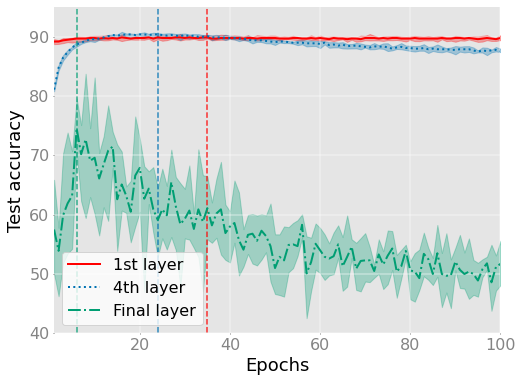}}}
\subfloat[Instance 40\%]{{\includegraphics[width=0.33\textwidth]{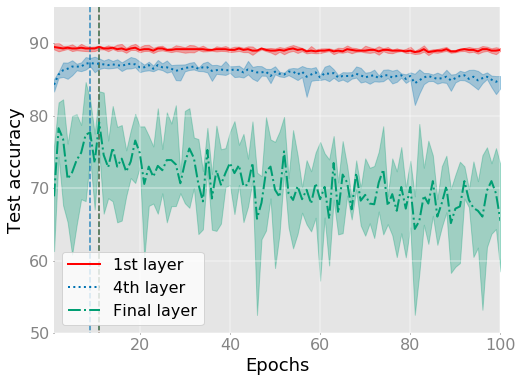}}}
\caption{We adopt LeNet as the model on Fashion-MNIST and evaluate the impact of noisy labels on the representations from the 1-st layer, the $4$-th layer, and the final layer. The curves present the mean of five runs and the best performances highlight with dotted vertical lines. Note that vertical lines are merged together for the 1-st layer and 4-th layer on Symmetric 50\%, and vertical lines of the 1-st layer and the final layer are merged together on Instance 40\%.}
\label{fig:5}
\vspace{-10px}
\end{figure}

% the green line (the final layer) decreases first, after several epochs, the blue line (the 33-th layer) begins to go down, and the performance of 17-th layer drops very late.

Figure \ref{fig:4} and Figure \ref{fig:5} demonstrate the impacts of noisy labels on different layers on CIFAR-100 and Fashion-MNIST, respectively. From Figure \ref{fig:4}, we can see that the drop of the green line (the final layer) is the largest, the blue line (the 33-th layer) has a gradual decline, and the orange line (the 17-th layer) is relatively stable during the training process. These observations are similar to those for CIFAR-10. The performance of 17-th layer in ResNet-34 is affected by noisy labels later and less than that of the 9-th layer in ResNet-18. It is because there are more layers after the 17-th layer in ResNet-34 than the 9-th layer. Similar trends are observable in Figure \ref{fig:5}. The first layer is nearly unaffected by noisy labels, and the performance of the final layer has a larger decline compared with the 4-th layer. The learning speeds of different layers are unapparent in LeNet, since there are only three hidden layers in LeNet, and the gradient of losses transfers much easier compared with deeper networks. Another reason may be the simplicity of patterns in Fashion-MNIST without image augmentation techniques, which leads the convolutional layers to learn fast.

\begin{table}[!h]
\vspace{-10px}
\centering
\caption{Comparison with state-of-the-art methods using ensemble two networks and semi-supervised learning on CIFAR-10 and CIFAR-100 with symmetric label noise from different levels. Baseline results are taken from \cite{Li2020DivideMix} and \cite{Liu2020ELR}. The highest results are reported for all the methods.}
\vspace{5pt}
\scalebox{1}{
{
\begin{tabular}{c | c | c | c | c | c | c }
\Xhline{2\arrayrulewidth}
	Dataset             &  \multicolumn{3}{c |}{CIFAR-10}                  & \multicolumn{3}{c}{CIFAR-100}                              \\ \hline
	Methods / Noise     & Sym-20\%       & Sym-50\%       & Sym-80\%       & Sym-20\%           & Sym-50\%          & Sym-80\%          \\ \hline
	CE                  & 87.2           & 80.9           & 65.8           & 58.1               & 47.5              & 23.6              \\
	MixUp               & 93.5           & 88.4           & 73.6           & 69.7               & 57.9              & 34.69             \\
	DivideMix*          & \textbf{96.1}  & 94.6           & 93.2           & 77.3               & 74.6              & 60.2              \\
	ELR+                & 95.8           & 94.8           & \textbf{93.3}  & 77.6               & 73.6              & 60.8              \\ 
    Ours (Semi)         & \textbf{96.1}  & \textbf{95.3}  & \textbf{93.3}  & \textbf{77.7}      & \textbf{74.9}     & \textbf{62.3}     \\
\Xhline{2\arrayrulewidth}
\end{tabular}
}}
\label{tab:cifar_sym_paper}
\end{table}

In the paper, we compare our results with baselines evaluated with a single network. In this section, we compare our method with state-of-the-art methods with ensemble two networks taken from the original papers \cite{Li2020DivideMix, Liu2020ELR}. We also adopt cross-entropy and MixUp \cite{Zhang2017MixUp} with a single network as baselines. From Table \ref{tab:cifar_sym_paper}, we can observe our results with a single network are comparable to results of baselines with ensemble two networks. Specifically, on CIFAR-100, our method outperforms state-of-the-art methods across all settings.

\end{document}